\documentclass[10pt,twocolumn,letterpaper]{article}

\usepackage{cvpr}      

\usepackage[english]{babel}
\usepackage{amssymb}
\usepackage{booktabs}
\usepackage{multicol}
\usepackage{multirow}
\usepackage{parskip}
\usepackage{amsmath}
\usepackage{caption}   
\usepackage{graphicx}  
\usepackage{color}
\usepackage{float}
\usepackage{colortbl}
\definecolor{pretty-blue}{RGB}{0, 113, 188}
\definecolor{icmlblue}{rgb}{0,0.08,0.45} 
\definecolor{linecolor1}{gray}{.95} 
\definecolor{linecolor}{gray}{.895} 

\usepackage[colorlinks=True,
            linkcolor=red,
            anchorcolor=blue,  
            pagebackref,
            urlcolor=black,
            citecolor=pretty-blue,
            ]{hyperref}

\usepackage[capitalize]{cleveref}
\crefname{section}{Sec.}{Secs.}
\Crefname{section}{Section}{Sections}
\Crefname{table}{Table}{Tables}
\crefname{table}{Tab.}{Tabs.}

\def\cvprPaperID{6504} 
\def\confName{CVPR}
\def\confYear{2023}

\begin{document}

\title{Structured Knowledge Distillation Towards Efficient and Compact \\ Multi-View 3D Object Detection}

\author{Linfeng Zhang$^{1*}$, Yukang Shi$^{2}$\thanks{The first two authors have equal contribution. This work was done during their intership in DIDI chuxing.}, Hung-Shuo Tai, Zhipeng Zhang, Yuan He, Ke Wang, Kaisheng Ma$^{1}\thanks{Corresponing author.}$\\
Tsinghua University$^{1}$, Xi'an Jiaotong University$^{2}$
}
\maketitle

\begin{abstract}
Detecting 3D objects from multi-view images is a fundamental problem in 3D computer vision. 
Recently, significant breakthrough has been made in multi-view 3D detection tasks.
However, the unprecedented detection performance of these vision BEV (bird's-eye-view) detection models is accompanied with enormous parameters and computation, which make them unaffordable on edge devices. To address this problem, in this paper, we propose  a structured knowledge distillation framework, aiming to improve the efficiency of modern vision-only BEV detection models.
The proposed framework mainly includes: (a) spatial-temporal distillation which distills teacher knowledge of information fusion from different timestamps and views, (b) BEV response distillation which distills teacher response to different pillars, and (c) weight-inheriting which solves the problem of inconsistent inputs between students and teacher in modern transformer architectures.
Experimental results show that our method leads to an average improvement of 2.16 mAP and 2.27 NDS on the nuScenes benchmark, outperforming multiple baselines by a large margin. 
%
\end{abstract}

\vspace{-0.1cm}
\section{Introduction}
\label{sec:intro}

Recently, bird's-eye-view (BEV) based multi-camera perception frameworks have greatly narrowed the performance gap with LiDAR based methods for 3D object detection tasks~\cite{bevdepth,bevformer,detr3d,fcos3d}.
For example, compared with state-of-the-art LiDAR methods, some recent works have obtained NDS scores within a 10\% margin~\cite{bevdepth,li2022bevstereo}.

Such vision-centric BEV frameworks usually involve two stages: single view feature extraction using backbone networks (convnets~\cite{liu2022convnet} or transformers~\cite{transformer}), and information fusion across multiple camera views and multiple timestamps using transformers~\cite{bevformer,liu2022petr,liu2022petrv2} or the lift-splat-shoot paradigm~\cite{huang2021bevdet,huang2022bevdet4d,bevdepth}. Once a spatial-temporal coherent feature representation is obtained in the unified BEV space, 3D object detection and semantic segmentation~\cite{bevformer,huang2021bevdet,bevdepth} can be done on the BEV feature map with high accuracy. 

\begin{figure}[t!]
    \centering
    \includegraphics[width=7.25cm]{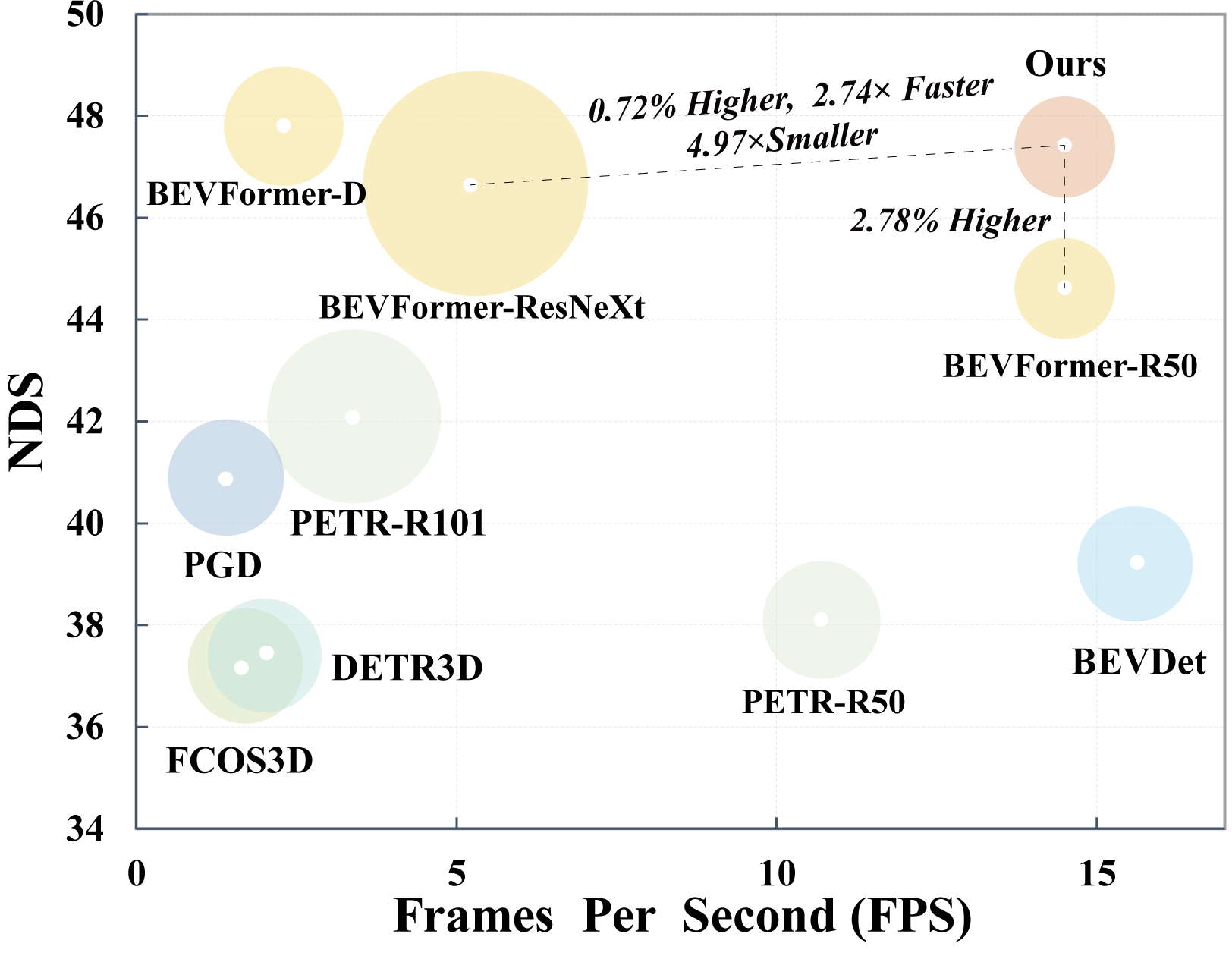}
    \vspace{-0.2cm}
    \caption{Experimental results on nuScenes. The area of circles indicate the number of parameters. Compared with BEVFormer-ResNeXt, our method achieves 0.72 NDS improvements, 2.74$\times$ acceleration and 4.97$\times$ compression. BEVFormer-ResNext\&R50 indicate teacher-2 and student-2 in Table~\ref{tab:stu_tea}, respectively.}
    \label{fig:performance}
    \vspace{-0.3cm}
\end{figure}

However, such performance improvements are achieved with a hefty computation overhead. For instance, 
the 120M parameters in BEVDet~\cite{huang2021bevdet} require more than 4 TFlops computation, which is almost 20$\times$ larger and 10$\times$ slower than CenterPoint~\cite{yin2021center}, a state-of-the-art LiDAR-based 3D detector.
Practical applications such as self-driving vehicles, usually have limited  computation budget but rather strict latency and accuracy requirements. Deployment of such visual BEV models onto edge devices requires a delicate balancing between low computation cost and high detection accuracy.
Compared with neural network pruning~\cite{deepcompression} and weight quantization~\cite{INQ,actication_quantization}, knowledge distillation (KD)~\cite{distill_hinton,model_compression} is more suited for striking such a balance.

Knowledge distillation is an effective model training technique that improves the performance of a lightweight student model by transferring the knowledge from a pre-trained but over-parameterized teacher model~\cite{model_compression,distill_hinton}.  At deployment time, only the lightweight student model is used.
While knowledge distillation has demonstrated great success in various 2D computer vision tasks, such as classification~\cite{selfdistillation}, object detection~\cite{kd_detection1,kd_detection2,detectiondistillation}, semantic segmentation~\cite{structured_kd,he2019knowledge}, and image generation~\cite{spkd_gan,omgd,distill_portable_gan,wkd}, 
the application of knowledge distillation on 3D computer vision, especially the camera-based multi-view 3D detection, has not been well-studied.
However, it is also brought to our attention that simply applying traditional knowledge distillation methods to 3D vision tasks usually leads to limited performance gains.

To address the aforementioned problems, this paper proposes a novel knowledge distillation framework for visual BEV detection models. We start with analyzing the challenges in the multi-view 3D detection task and then propose the corresponding solution as follows:
\vspace{-0.4cm}
\paragraph{Information fusion from multiple positions:} 
In multi-view 3D detection, the detector takes input from multiple cameras across different timestamps to identify objects.
Hence, the student should be able to learn not only the information from single images but also how to fuse and leverage the information from multiple spatial/temporal positions.
To tackle this challenge, we propose spatial-temporal distillation, which improves student performance by allowing it to learn the semantic correspondence between inputs in different spatial (\emph{i.e.,} view) and temporal positions from their teachers.
Moreover, we also propose BEV response distillation, which aims to distill teacher response to different positions/pillars in the BEV feature map, which contains high level information on object localization.

\vspace{-0.4cm}
\paragraph{Discrepancies between the inputs:} The state-of-the-art BEV 3D detectors usually employ a DETR-like architecture, which utilizes self-attention and cross-attention layers for information fusion~\cite{detr,detr3d}. Different from traditional convolutional detectors, the input information of DETR-style detectors contains not only images but also trainable queries and positional encodings. 
Without explicit constraints, student and teacher models could have learned different positional encodings and queries after training. Knowledge distillation will be hindered by such discrepancies~\cite{beyer2022knowledge}.
To address this problem, we propose a weight-inheriting scheme which fixes the positional encodings and BEV queries of the student model to the corresponding values in the teacher detector. In this way, the student detector will benefit from the pre-trained weights of the teacher detector directly.
Surprisingly, we find that even without applying any knowledge distillation losses, simply using the weight-inheriting scheme can significantly improve the performance of knowledge distillation for this task.

Without loss of generality, we conduct extensive experiments on the nuScenes dataset~\cite{caesar2020nuscenes} using a representative and state-of-the-art BEVFormer model architecture~\cite{bevformer}. 
On average, 2.16 mAP and 2.27 NDS improvements can be observed across three different student-teacher settings, demonstrating the effectiveness of our proposed knowledge distillation framework. Compared with multiple baseline methods~\cite{distill_hinton,attentiondistillation,kd_comprehensive,relation_detection,relational_kd2,kd_variational,detectiondistillation,DBLP:conf/cvpr/Guo00W0X021}, our method outperforms them all by a large margin.

In summary, our contributions include:

\noindent
(\textbf{1}) We propose a novel spatial-temporal distillation scheme which enables the student detector to learn teacher knowledge on how to fuse information from different camera views and timestamps. 

\noindent
(\textbf{2}) BEV response distillation is proposed to distill teacher response to different BEV pillars, which transfers teacher knowledge on object localization to the student.

\noindent
(\textbf{3}) We identify the problem of inconsistent inputs in knowledge distillation on DETR-style detectors and propose a weight-inheriting scheme to solve it.

\noindent
(\textbf{4}) Extensive experiments on nuScenes demonstrate the effectiveness of our method. On average, \textbf{2.16} mAP and \textbf{2.27} NDS improvements can be obtained compared with the student without knowledge distillation. Source code and models will be released to benefit the community.

\begin{figure*}
    \centering
    \includegraphics[width=0.95\linewidth]{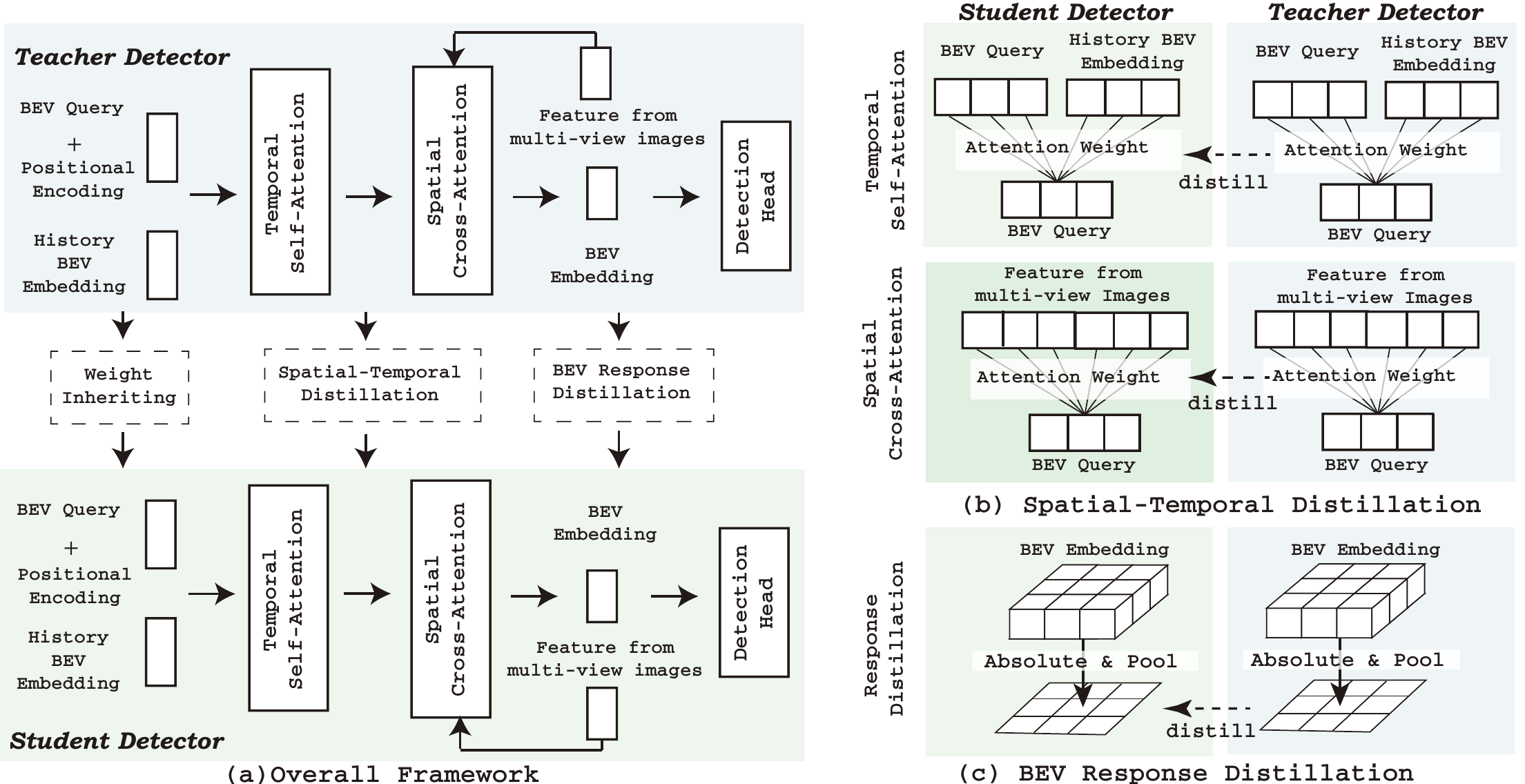}
    \vspace{-0.25cm}
    \caption{The overall framework and details of our method. 
    (a) The proposed knowledge distillation methods mainly include weight-inheriting, spatial-temporal distillation, and BEV response distillation. Weight-inheriting fixes the parameters of BEV queries and positional encoding to their value in the pre-trained teacher detector during the whole training period to guarantee that students and teachers have the same inputs. (b) Spatial-temporal distillation aims to improve student performance on information fusion of images from multiple views and timestamps by transferring teacher knowledge in the attention weights in the temporal self-attention and spatial cross-attention layers. (c) BEV response distillation first computes the response of different positions in BEV map and then distills it to the students.
    }
    \vspace{-0.4cm}
    \label{fig:main_figure}
\end{figure*}
\vspace{-0.1cm}
\section{Related Work}
\label{sec:related_work}
\subsection{Knowledge Distillation}
Knowledge distillation, which aims to transfer knowledge from a cumbersome teacher model to a lightweight student model, has become one of the most popular model training techniques in both model compression and model performance boosting~\cite{distill_hinton,model_compression}. 
Knowledge distillation is originally proposed to train the student model to mimic the output probability distribution from teachers for image classification tasks~\cite{distill_hinton,deepmutuallearning}.
In addition to model outputs, intermediate features~\cite{fitnets,kd_crd}, attentions~\cite{attentiondistillation}, as well as relations~\cite{relational_kd,relational_kd2,fsp_kd,kd_cc}, all have been utilized for knowledge distillation.
Nowadays, knowledge distillation has successfully swept a wide range of applications, such as object detection~\cite{detectiondistillation,kd_detection4,kd_detection3}, semantic segmentation~\cite{structured_kd,he2019knowledge}, image generation~\cite{wkd}, self-supervised learning~\cite{kd_self1,kd_self2}, vision model pretraining~\cite{simclr,simclr2,moco}, language models~\cite{kd_bert1,kd_bert2}, data augmentation~\cite{labelrefine}, model robustness~\cite{kd_defense,auxiliarytraining} and so on.

\vspace{-0.4cm}
\paragraph{KD for 3D Detection}
Following the success of knowledge distillation on 2D tasks, a few works have been proposed to apply knowledge distillation to 3D object detection tasks.
For example, in the point cloud detection domain,
Zhang~\emph{et al.\@}~\cite{zhang2022pointdistiller} distills local information extracted by dynamic graph convolutions,
Cho~\emph{et al.\@}~\cite{cho2022itkd} distills the knowledge compressed by an AutoEncoder,
Yang~\emph{et~al.\@}~\cite{yang2022towards} propose to perform knowledge distillation only on the positions with high teacher classification responses,
Hou~\emph{et~al.\@}~\cite{hou2022point} propose to compress 3D segmentation by distilling the knowledge in both points and voxels,
Huang~\emph{et al.\@}~\cite{huang2022label} proposed label-guided auxiliary training which generates pseudo teacher features with label information.
In the vision based 3D detection domain,
Chong~\emph{et al.\@}~\cite{chong2022monodistill} improves the efficiency of monocular 3D detection with relation-based knowledge distillation.
Guo~\emph{et~al.\@} propose to learn stereo-based students from a LiDAR-based teacher~\cite{guo2021liga}. Sautier~\emph{et~al.\@} propose image-to-LiDAR self-supervised distillation which leverages the information of an image-based detector to improve the performance of 3D detector~\cite{sautier2022image}. 
Despite much success in these prior works, applying knowledge distillation to the modern multi-camera BEV detection paradigm has been rarely explored. We hope our attempts can inspire future research in this domain.

\vspace{-0.1cm}
\subsection{Camera-based 3D Detection}
Detecting 3D objects from images is one of the most challenging problems in 3D computer vision. 
With the help of modern deep learning techniques, much progress has been made for monocular 3D object detection~\cite{3d_object_1,3d_object_2,3d_object_3,fcos3d}.
However, monocular 3D object detection often suffer from truncation and occlusion problems, which are quite challenging without the help of additional sensors.
Detecting 3D objects using multi-cameras in the Birds'-Eye-View (BEV) thus have become quite popular recently.
Huang~\emph{et~al.\@} propose BEVDet which encodes the feature of single views and then projects them to BEV space~\cite{huang2022bevdet4d,huang2021bevdet}.
Wang~\emph{et al.} propose Detr3D which manipulates predictions directly in 3D space by using 3D object queries to index the features of multi-view images~\cite{detr3d}.
PETR and PETRv2 are proposed to produce 3D position-aware features by encoding the position information of 3D coordinates into image features~\cite{liu2022petr,liu2022petrv2}.
Li~\emph{et al.} propose BEVFormer, which employs spatial cross-attention and temporal self-attention to merge the information of features in different spatial and temporal positions~\cite{bevformer}.
Then, BEVDepth is proposed to perform explicit depth supervision with encoded intrinsic and extrinsic parameters~\cite{bevdepth}.
Observing the fact that the optimal temporal difference between views varies significantly for different pixels and depths,
Park~\emph{et~al.} propose SOLOFusion which employs both short-term and long-term temporal stereo for depth estimation~\cite{park2022time}.
Li~\emph{et al.} propose BEVStereo which dynamically selects the scale of matching candidates to reduce the computation overhead~\cite{li2022bevstereo}. 
Following such trend, we focus on designing effective knowledge distillation strategies to push the envelope of such multi-camera BEV object detection paradigm.
\vspace{-0.3cm}
\section{Methodology}
\label{sec:methodology}

\subsection{Preliminary}
Without loss of generality, we conduct our experiments on top of the BEVFormer model, which is a performant and representative multi-view 3D object detection architecture. To recap, BEVFormer consists of four stages, including feature extraction from single images, temporal information fusion, spatial information fusion, and prediction.

\vspace{-0.35cm}
\paragraph{(I) Feature Extraction of Single Images:} In multi-view 3D detection, at timestamp $t$, the input image set can be denoted as $\mathcal{X}^{(t)}=\{x_1^{(t)}, x_2^{(t)}, ..., x_v^{(t)}\}$, where $v$ denotes the number of views.
BEVFormer firstly encodes the feature of each single image with a convolutional 2D backbone $f_{\text{2D}}$, which can be formulated as $F^{(t)}=f_{\text{2D}}(\mathcal{X}^{(t)})$. These features are then fed into spatial cross-attention in Stage~III for multi-view feature fusion.

\vspace{-0.35cm}
\paragraph{(II) Temporal Information Fusion:} Temporal self-attention is utilized to fuse the information between the current input images and the historical images.
The input of temporal self-attention layers includes the predefined trainable BEV queries with positional encoding and the previous BEV embedding at timestamp $t-1$, which can be formulated as 
$Q^{\text{BEV}}$ and $E_{\text{BEV}}^{(t-1)}$, respectively.
Then, the computation of temporal self-attention can be written as
\begin{equation}
    \begin{aligned} 
    E'^{(t)}_{\text{BEV}} &= \text{TSA}\left(Q^{\text{BEV}}_p, \{Q^{\text{BEV}}, E_{\text{BEV}}^{(t-1)}\}\right)
    \\&=\mathop{\sum\nolimits_{V\in\{Q^{\text{BEV}}, E_{\text{BEV}}^{(t-1)}\}}} \text{DeformAttn}(Q_p^{\text{BEV}}, p, V),
    \end{aligned}
\end{equation}
where DeformAttn indicates the deformable attention layers~\cite{zhu2020deformable} and $Q_p^{\text{BEV}}$ denotes 
the BEV query located at the position $p$. TSA and $E'^{(t)}_{\text{BEV}}$ indicate temporal self-attention and its outputs, respectively.

\vspace{-0.35cm}
\paragraph{(III) Spatial Information Fusion:} In the stage of spatial information fusion, BEVFormer samples $N_{\text{ref}}$ 3D reference points from each pillar, and then projects them to 2D views.
Then, spatial cross-attention is utilized to fuse the BEV embedding output by temporal information fusion with the reference points, which can be formulated as 
\begin{equation}
        E^{(t)}_{\text{BEV}} =\frac{1}{|v_{\text{hit}}|}\mathop{\sum_{i\in v_{\text{hit}}}}\mathop{\sum_{j=1}^{N_{\text{ref}}}} \text{DeformAttn}\left(E'^{(t)}_{\text{BEV}}, \mathcal{P}(p,i,j), F^{(t)}\right),
\end{equation}
where $v_{\text{hit}}$ indicates the number of views that contain the projection of the 3D reference points. 
$\mathcal{P}(p,i,j)$ is the projection function to get the $j$-th reference point on the $i$-th view image. $F^{(t)}$ indicates the feature of single images computed in Stage~I.

\vspace{-0.35cm}
\paragraph{(IV) Prediction} In this stage, BEVFormer predicts the positions, dimensions, headings, and categories of objects based on the two inputs, including the output of spatial cross-attention and a set of object queries, which can be denoted as  $Q^{\text{Object}}$ and $E^{(t)}_{\text{BEV}}$, respectively.
Its computation can be formulated as 
\begin{equation}
    \begin{aligned}
        \text{B}, \text{P} = \text{Detection Head}(E^{(t)}_{\text{BEV}}, Q^{\text{Object}}), 
    \end{aligned}
\end{equation}
where ``B'' and ``P'' indicate the predicted bounding boxes and the corresponding probability distribution.

\subsection{Structured Knowledge Distillation\label{sec:method}}
In this subsection, we introduce the proposed knowledge distillation based on the above four stages in BEVFormer.
Note that the Stage~I (2D convolutional feature extraction) and Stage~IV (prediction) in BEVFormer share quite some similarities with common 2D detectors. Successful attempts have been made to apply knowledge distillation onto these stages~\cite{kd_detection1,detectiondistillation}. Thus in this paper, we focus on the Stage~II and Stage~III, which are critical for multi-view 3D detection but rarely explored for knowledge distillation.
In particular, our method can be divided into the following three folds.


\vspace{-0.35cm}
\paragraph{Spatial-Temporal Knowledge Distillation} 
In Stage~II and Stage~III, BEVFormer first integrates the BEV queries with the BEV embeddings at the previous timestamp for temporal information fusion, and then fuses the information from different image views for spatial information fusion.
Deformable attention layers are utilized during the two processes. 
Recall that the computation of attention weights in  deformable attention layers is obtained by a linear projection over queries followed with a softmax function, which 
can be formulated as 
\begin{equation}
    \mathcal{A}(\mathbf{Q}) =  \text{softmax}(\mathbf{W}\mathbf{Q}),
\end{equation}
where $\mathbf{Q}$ and $\mathbf{W}$ indicate the queries and the trainable parameters in the linear projection layer, respectively. 
In temporal self-attention, $\mathbf{Q}$ indicates the BEV query $Q^{\text{BEV}}$. And the obtained attention weights are utilized to fuse information from $Q^{\text{BEV}}$ and the historical BEV embedding $E_{\text{BEV}}^{(t-1)}$. Hence, the attention weights here show the temporal relation between the information of the current inputs and the previous input.
By distilling them, the student is allowed to learn how to fuse temporal information from the teacher detector.
In spatial cross-attention, $\mathbf{Q}$ indicates the output of temporal self-attention $E'^{(t)}_{\text{BEV}}$. And the obtained attention weights are utilized to fuse the information from the reference points in the multi-view images. 
Hence, distilling the attention weights here enables the student to learn how to fuse spatial information from the teacher detector.
Concretely, we can denote the attention weights in temporal self-attention and temporal self-attention as $\mathbf{A}^{\text{temporal}}$ and $\mathbf{A}^{\text{spatial}}$, respectively, which can be formulated as  
\begin{equation}
  \begin{aligned}
  \mathbf{A}^{\text{temporal}} &= \mathcal{A}(Q^{\text{BEV}}), \text{~and} \\
  \mathbf{A}^{\text{spatial}} &= \mathcal{A}(E'^{(t)}_{\text{BEV}}) \text{,~~respectively.}
  \end{aligned}
\end{equation}
Then, by distinguishing the student and teacher detector with the scripts $\mathcal{S}$ and $\mathcal{T}$ respectively, the proposed spatial-temporal attention can be formulated as
\begin{equation}
    \mathcal{L}_{\text{spatial-temp}} = \| \mathbf{A}^{\text{temporal}}_{\mathcal{S}} - \mathbf{A}^{\text{temporal}}_{\mathcal{T}} \|^2 + \| \mathbf{A}^{\text{spatial}}_{\mathcal{S}} - \mathbf{A}^{\text{spatial}}_{\mathcal{T}} \|^2.
\end{equation}

\vspace{-0.35cm}
\paragraph{BEV Response Distillation}
Besides distilling teacher knowledge on the fusion of the information from different timestamps and views, we also propose BEV response distillation to distill teacher responses to different object queries, which correspond to different pillars in 3D space.
In this paper, we define the BEV response as the average score across the channel dimension on the absolute value of BEV embedding, which can be written as 
\begin{equation}
    \mathcal{R}(E_{\text{BEV}(i,j)}) = \sum\nolimits_{j=1}^C \frac{1}{C} |E_{\text{BEV}(i,j)}|,
\end{equation}
where $C$ denotes the number of channels. The scripts $(i,j)$ denotes the value on the i$_{th}$ BEV query (\emph{i.e.,} pillar) of the j$_{th}$ channel. As pointed out by abundant research~\cite{detectiondistillation,attentiondistillation,zhang2022region}, the response of features demonstrates the importance of their corresponding spatial positions.
Hence, by distilling the BEV response from the teacher, the student model can better correlate between the learned semantic features and the potential object spatial occupancies. An example of BEV response is visualized in~\cref{fig:bev_vis}.
An L2 loss is adopted for BEV response distillation:
\begin{equation}
    \mathcal{L}_{\text{response}} = \|\mathcal{R}(E_{\text{BEV})}^\mathcal{S} - \mathcal{R}(E_{\text{BEV}}^\mathcal{T})\|^2, 
\end{equation}
where $\mathcal{S}$ and $\mathcal{T}$ denote the student detector and the teacher detector, respectively.
Based on the above notations, the overall training loss of the detector $\mathcal{L}$ becomes:
\begin{equation}
\mathcal{L} = \mathcal{L}_{\text{original}} + \lambda \cdot ( \mathcal{L}_{\text{spatial-temp}} + \mathcal{L}_{\text{response}}),
\end{equation}
where $L_{\text{original}}$ indicates the original training loss of BEVFormer. $\lambda$ is a hyper-parameter to balance the magnitudes of knowledge distillation loss, which is set to 1$\times$10$^{-2}$ in all the experiments.
Please refer to the supplementary material for its sensitivity study.

\vspace{-0.35cm}
\paragraph{Weight-Inheriting}
Convnets-based detectors usually only require images as input. But modern transformer-based detection models require additional learned queries and positional encodings as input. The teacher and the student model tend to have different query and positional encoding values after training converges.
Intuitively, knowledge distillation works by aligning the output of the student with the teacher given the same input.
Such paradigm is likely to fail for transformer-based detectors, as the teacher and student can have different learned queries and position encodings. The discrepancies between the transformer inputs must be resolved to make the underlying assumptions of knowledge distillation hold true.

Hence, in this paper, we propose a weight-inheriting scheme that fixes the value of the BEV queries and positional encoding in the student with their values from the teacher detector \emph{during the whole training period}. Hence, the student detector can have consistent inputs with its teacher detector. 
Surprisingly, we find that simply performing this weight-inheriting scheme can make a significant difference in the effectiveness of knowledge distillation, which has been discussed in the ablation study.




%
\vspace{0.4cm}
\begin{table}
    \centering
      \caption{Student-teacher settings in our experiments. Please refer to the supplementary material for more details.}
    \vspace{-0.3cm}
      \resizebox{\linewidth}{!}{
    \begin{tabular}{c|clccccc}
    \toprule
        Model& FPS&Params&2D Backbone & BEV Query & Decoder Depth\\
        \midrule
        Student-1& 14.5&40.45&ResNet50&(150, 150)& 3 \\
        Teacher-1& 10.2&56.57&ResNet101& (150, 150) & 3\\ \midrule
        Student-2& 14.5&40.45&ResNet50&(150, 150)& 3 \\
        Teacher-2& 5.3&201.20&ResNeXt-Large& (150, 150) & 3\\ \midrule
        Student-3& 5.2&47.56&ResNet50&(200, 200)& 6 \\
        Teacher-3& 3.5&65.93&ResNet101& (200, 200) & 6\\ \bottomrule

    \end{tabular}}

    \label{tab:stu_tea}
    \vspace{-1em}
\end{table}

\begin{table*}[t]
    \caption{Comparison with other knowledge distillation methods on the nuScenes~\cite{caesar2020nuscenes} dataset with BEVFormer. Note that a higher mAP and NDS, as well as a lower ATE, ASE, AOE, and AAE indicate better performance. Params: the number of parameters (M). FPS: Frame per second. FPS is measured with one A100 GPU. Please refer to ~\cite{caesar2020nuscenes} for detailed metrics definitions.\label{tab:nuscene}}
    \vspace{-0.25cm}
    \begin{center}
  \resizebox{\linewidth}{!}{\begin{tabular}{lccl|cccccccccccccc}
    \toprule
        Backbone&FPS&Params& KD Method  &mAP($\uparrow$)&NDS($\uparrow$)&mATE($\downarrow$)&mASE($\downarrow$)&mAOE($\downarrow$)&mAVE($\downarrow$)&mAAE($\downarrow$)\\
        \midrule
        ResNet101&10.2&56.57& Teacher w/o KD&36.31&47.49&69.21&28.16&46.08&43.87&19.32& \\
        \midrule
        \multirow{10}{*}{ResNet50}&\multirow{10}{*}{14.5}&\multirow{10}{*}{40.45}& Student w/o KD&33.56&44.61&71.41&28.65&54.17&46.44&21.03 \\
        &&& + Hinton~\emph{et al.}&        33.57&45.23&71.17&28.50&49.04&46.52&20.33 \\
        &&& + Zagoruyko~\emph{et al.}& 
33.68&45.69&70.13&\textbf{27.74}&47.87&45.45&20.26\\
        &&& + Heo~\emph{et al.}&         33.87&45.82&69.92&27.79&47.78&45.55&20.09\\
&&& + Park~\emph{et al.}&33.77&45.87&70.88&27.78&48.18&43.47&19.83 \\
        &&& + Pung~\emph{et al.}&34.01&45.36&71.21&28.06&50.49&45.77&20.88 \\
        &&& + Ahn~\emph{et al.}& 34.11&46.36&70.69&28.02&46.16&42.09&20.04\\
&&& + Zhang~\emph{et al.}& 34.25&46.34&70.84&28.44&47.06&\textbf{41.68}&19.82\\

&&& + Guo~\emph{et al.}& 34.10&46.22&70.39&28.39&46.75&42.52&20.22\\
&&& \textbf{+ Ours}\cellcolor{linecolor}&\textbf{34.91}\cellcolor{linecolor}&\textbf{46.87}\cellcolor{linecolor}&\textbf{69.77}\cellcolor{linecolor}&28.07\cellcolor{linecolor}&\textbf{46.31}\cellcolor{linecolor}&42.23\cellcolor{linecolor}&\textbf{19.43}\cellcolor{linecolor} \\
         \midrule
        ResNeXt-Large&5.3&201.2& Teacher w/o KD&37.69&46.67&70.44&28.52&56.89&45.81&20.12 \\
        \midrule
        \multirow{10}{*}{ResNet50}& \multirow{10}{*}{14.5}&\multirow{10}{*}{40.45} & Student w/o KD&33.56&44.61&71.41&28.65&54.17&46.44&21.03 \\
        &&& + Hinton~\emph{et al.}&33.84&45.68&72.72&28.16&46.54&44.50&20.48\\
        &&& + Zagoruyko~\emph{et al.}&34.10&46.26&70.99&28.24&46.12&42.45&20.05\\
        &&& + Heo~\emph{et al.}& 34.30&46.50&70.36&27.94&\textbf{44.78}&43.06&20.39\\
        &&& + Park~\emph{et al.}&33.98&46.40&71.82&28.07&45.84&39.86&20.24 \\
        &&& + Pung~\emph{et al.}&34.23&46.23&70.05&28.32&47.33&43.13&20.04 \\
        &&& + Ahn~\emph{et al.}& 34.16&46.25&70.37&28.08&46.43&42.73&20.66\\
        &&& + Zhang~\emph{et al.}&34.56&46.61&70.11&28.01&46.00&42.39&20.14 \\
        &&& + Guo~\emph{et al.}&34.35&46.06&69.92&\textbf{27.79}&47.78&45.55&20.09 \\
        &&& \textbf{+ Ours}\cellcolor{linecolor}& \textbf{35.58}\cellcolor{linecolor}&\textbf{47.39}\cellcolor{linecolor}&\textbf{68.97}\cellcolor{linecolor}&28.25\cellcolor{linecolor}&48.06\cellcolor{linecolor}&\textbf{39.79}\cellcolor{linecolor}&\textbf{18.93}\cellcolor{linecolor}\\
        \midrule
        ResNet101&3.5&65.93& Teacher w/o KD&41.01&51.88&67.45&27.36&34.92&37.57&18.97 \\
        \midrule
        \multirow{10}{*}{ResNet50}&\multirow{10}{*}{5.2}&\multirow{10}{*}{47.56}& Student w/o KD&35.77&46.74&73.61&28.26&45.85&43.79&19.94 \\
        &&& + Hinton~\emph{et al.}&35.89&46.93&73.45&\textbf{28.02}&45.46&43.66&19.58\\
        &&& + Zagoruyko~\emph{et al.}&35.98&46.98&73.30&28.22&45.32&43.68&19.60\\ 
        &&& + Heo~\emph{et al.}&36.23&47.16&73.09&28.18&45.28&43.34&19.69\\ 
        &&& + Park~\emph{et al.}& 36.30&47.18&72.94&28.17&45.48&43.43&19.64\\
        &&& + Pung~\emph{et al.}&36.42&47.26&72.96&28.23&45.48&43.37&19.51\\ 
        &&& + Ahn~\emph{et al.}& 36.38&47.20&73.02&28.25&45.51&43.50&19.60\\
        &&& + Zhang~\emph{et al.}&36.64&47.38&73.12&28.15&\textbf{45.28}&43.11&19.53 \\
        &&& + Guo~\emph{et al.}&36.77&47.40&73.14&28.25&45.34&43.43&19.74\\ 
        &&& \textbf{+ Ours}\cellcolor{linecolor}&\textbf{38.88}\cellcolor{linecolor}&\textbf{48.52}\cellcolor{linecolor}&\textbf{71.53}\cellcolor{linecolor}&28.24\cellcolor{linecolor}&47.34\cellcolor{linecolor}&\textbf{42.91}\cellcolor{linecolor}&\textbf{19.17}\cellcolor{linecolor} \\
         \bottomrule 
    \end{tabular}}
    \vspace{-0.2cm}
    \end{center}
\end{table*}

\vspace{-0.4cm}
\section{Experiment}
\vspace{-0.1cm}
\label{sec:experiment}
\subsection{Experimental Setting}
\paragraph{Dataset:} The nuScenes dataset is a large-scale autonomous driving dataset, which has 3D bounding boxes for 1000 scenes collected from six cameras~\cite{caesar2020nuscenes}. The scenes are officially split into 700, 150, and 150 scenes for training, validation, and testing, respectively, including 1.4 million annotated 3D bounding boxes belonging to 10 classes.


\vspace{-0.40cm}
\paragraph{Model Architecture:} BEVFormer models of different sizes are utilized as the student and teacher detectors in our experiments. As shown in Table~\ref{tab:stu_tea}, We mainly reduce the model size by using fewer BEV queries and smaller 2D backbones. Please refer to the supplementary material for more details on the models and training settings.

\vspace{-0.40cm}
\paragraph{Comparison Method:}
Eight previous knowledge distillation methods have been utilized for comparison. Six of the comparison methods are feature-based knowledge distillation for general vision tasks, including methods from Hinton~\emph{et al.}~\cite{distill_hinton}, Zagoruyko~\emph{et al.}~\cite{attentiondistillation}, Heo~\emph{et al.}~\cite{kd_comprehensive}, Park~\emph{et~al.}~\cite{relational_kd}, Pung~\emph{et al.}~\cite{relational_kd2} and Ahn~\emph{et al.}~\cite{kd_variational}. 
We adopt them to multi-view detection by using them to distill the features of both the 2D backbone and self-attention layers. Besides, two of the comparison methods are proposed for 2D object detection, including methods from Zhang~\emph{et~al}~\cite{detectiondistillation} and Guo~\emph{et~al.}~\cite{DBLP:conf/cvpr/Guo00W0X021}. Please refer to the supplementary material on our detailed implementation.


\subsection{Experiment Results}
\vspace{-0.1cm}
Experimental results of our method and eight previous knowledge distillation methods in three different student-teacher settings are shown in Table~\ref{tab:nuscene}. It is observed that: (i)~On average, 2.16 mAP and 2.27 NDS improvements can be observed with our method in the three student-teacher settings, which are 1.26 mAP and 0.80 NDS higher than the second-best knowledge distillation methods. (ii) In all three student-teacher settings, our method leads to performance improvements in terms of most of the performance metrics, including mAP, NDS, mATE, mATE, mASE, mAOE, mAVE, and mAAE, indicating that our method benefits students in estimating the translation, scale, orientation, velocity and attributes of the objects. (iii) The performance of our method in different categories is shown in Table~\ref{tab:class}. It is observed that our method leads to consistent improvements in most of the categories. (iv) The first student achieves 0.67 higher mAP than the second student, indicating that our method benefits from a strong teacher.


\begin{table}[t]
  \caption{Average precision in different classes on nuScenes. ``KD'' indicates whether our method is applied. Experiments of the three groups are conducted with student-teacher settings in Table~\ref{tab:stu_tea}. \label{tab:class}}
  \vspace{-0.3cm}
  \begin{center}
    \resizebox{\linewidth}{!}{\setlength{\tabcolsep}{0.50mm}{\begin{tabular}{c|ccccccccccccccccc}
      \toprule
         KD  &~~~Car~~~&Truck&~~~Bus~~~&Trailer&Con.Veh.& Pedest.&Motor.&Bicycle&Barrier&Tra.Cone\\
           \midrule
          $\times$&54.3&26.0&32.3&8.9&7.4&41.8&31.8&28.2&53.8&51.0\\
          $\checkmark$&55.1&27.4&34.2&10.1&6.8&43.4&34.2&31.1&53.9&52.9\\
          \midrule[0.1pt]
          $\times$&54.3&26.0&32.3&8.9&7.4&41.8&31.8&28.2&53.8&51.0\\
          $\checkmark$&56.5&29.3&37.5&13.3&10.3&45.6&34.4&34.4&43.8&50.8\\
          \midrule[0.1pt]
          $\times$&55.8&28.7&35.0&9.7&6.5&46.6&37.5&37.3&54.6&46.0\\
          $\checkmark$&58.7&33.1&36.2&12.8&10.1&47.4&40.4&40.8&57.6&51.9\\
           \bottomrule 
      \end{tabular}}}
      \end{center}
      \vspace{-1.3em}
  \end{table}

\begin{table*}[h]
  \caption{Ablation study of different modules in our method. ``Spatial-Temporal'', ``BEV Response'', ``Weight-Inherit'' indicates spatial-temporal distillation, BEV response distillation, and the weight-inheriting scheme, respectively.
  \label{tab:ablation}}
      \vspace{-0.25cm}
  \begin{center}
  \small
\resizebox{\linewidth}{!}{\begin{tabular}{cccccccccccccccccc}
  \toprule
       \multicolumn{3}{c}{Modules in Our Method} &  &\multirow{2}{*}{mAP($\uparrow$)}&\multirow{2}{*}{NDS($\uparrow$)}&\multirow{2}{*}{mATE($\downarrow$)}&\multirow{2}{*}{mASE($\downarrow$)}&\multirow{2}{*}{mAOE($\downarrow$)}&\multirow{2}{*}{mAVE($\downarrow$)}&\multirow{2}{*}{mAAE($\downarrow$)}\\
       \cmidrule{1-3}
       {{\small Spatial-Temporal}} &{{\small BEV Response}}& {{\small Weight-Inherit}} \\
       \midrule
      $\times$&$\times$&$\times$&&33.56&44.61&71.41&28.65&54.17&46.44&21.03\\
      $\times$&$\times$&$\checkmark$&&34.52&46.60&70.97&28.05&46.00&41.74&19.86\\
      $\times$&$\checkmark$&$\checkmark$&&34.99&47.17&70.22&27.75&46.58&39.47&19.24\\
      $\checkmark$&$\times$&$\checkmark$&&34.91&47.02&70.62&27.90&47.26&39.60&18.93 \\
      $\checkmark$&$\checkmark$&$\times$&&35.00&46.68&71.07&28.43&46.09&42.46&20.19& \\
      $\checkmark$&$\checkmark$&$\checkmark$&& 35.58&47.39&68.97&28.25&48.06&39.79&18.93\\
       \bottomrule 
  \end{tabular}}
  \end{center}
\end{table*}


\section{Discussion}
\label{sec:discussion}

\subsection{Ablation Study}
The proposed knowledge distillation methods mainly have three modules, including spatial-temporal distillation, the BEV embedding distillation, and the weight-inheriting scheme. Table~\ref{tab:nuscene} gives the ablation study of the three modules. It is observed that: (i) By simply using the weight-inheriting scheme without applying any knowledge distillation loss, 0.96 mAP and 1.99 NDS improvements can be obtained, indicating that the student detector can benefit from using the pre-trained weights from teachers on the BEV queries and positional encoding.
(ii) By applying BEV response distillation and weight-inheriting, 1.43 mAP and 2.56 NDS improvements can be observed, which are 0.47 and 0.57 higher than only using weight-inheriting, indicating BEV response distillation can successfully transfer teacher knowledge to the student.
(iii) 1.35 mAP and 2.41 NDS improvements can be obtained by using spatial-temporal distillation and weight-inheriting, which are 0.39 and 0.42 higher than only using weight-inheriting, indicating spatial-temporal distillation allows the student to learn how to fuse information from different timestamps and views from its teacher.
(iv) By combining the three modules together, 1.67 mAP and 2.58 NDS improvements can be obtained, which demonstrates that the benefits of spatial-temporal distillation and BEV response distillation are orthogonal. (v) By only using the two knowledge distillation while disabling the weight-inheriting scheme, 1.44 mAP and 2.07 NDS improvements can be observed, which are 0.58 and 0.71 lower than performing knowledge distillation with weight-inheriting, indicating weight-inheriting is also indispensable even if knowledge distillation losses are applied.
In summary, these experimental results demonstrate that the three modules in our method have their own effectiveness and their merits are orthogonal.
\vspace{-0.40cm}
\paragraph{Ablation on Weight-Inheriting}
To facilitate the training of the student model, some previous knowledge distillation methods propose initializing the parameters of the student with the parameters of the teacher (\emph{i.e.,}  initialization scheme), which sometimes leads to slight performance improvements. In contrast, the proposed weight-inheriting scheme in this paper not only initializes the parameters of BEV queries and positional encoding with their value from the teacher but also freezes them during the whole training period (\emph{i.e.,} weight-inheriting scheme). To study their difference, we have conducted several experiments and found that (i) By using the initialization scheme, after the training of the student, the parameters of BEV queries and positional encoding in the student are totally different from them in the teacher, indicating the inconsistency problem between the students and the teachers in knowledge distillation still exist. (ii) Experimental results show that by only using the traditional initialization scheme, the student detector (student-1 in Table~\ref{tab:stu_tea}) achieves 33.63 mAP and 44.70 NDS, which are 0.7 and 0.9 higher than the baseline, but still 0.89 and 1.90 lower than the weight-inheriting scheme. These observations indicate that using such a weight-inheriting scheme which exactly guarantees the consistency between the inputs of the students and teachers is indispensable.  



\begin{figure}[t!]
    \centering
    \includegraphics[width=\linewidth]{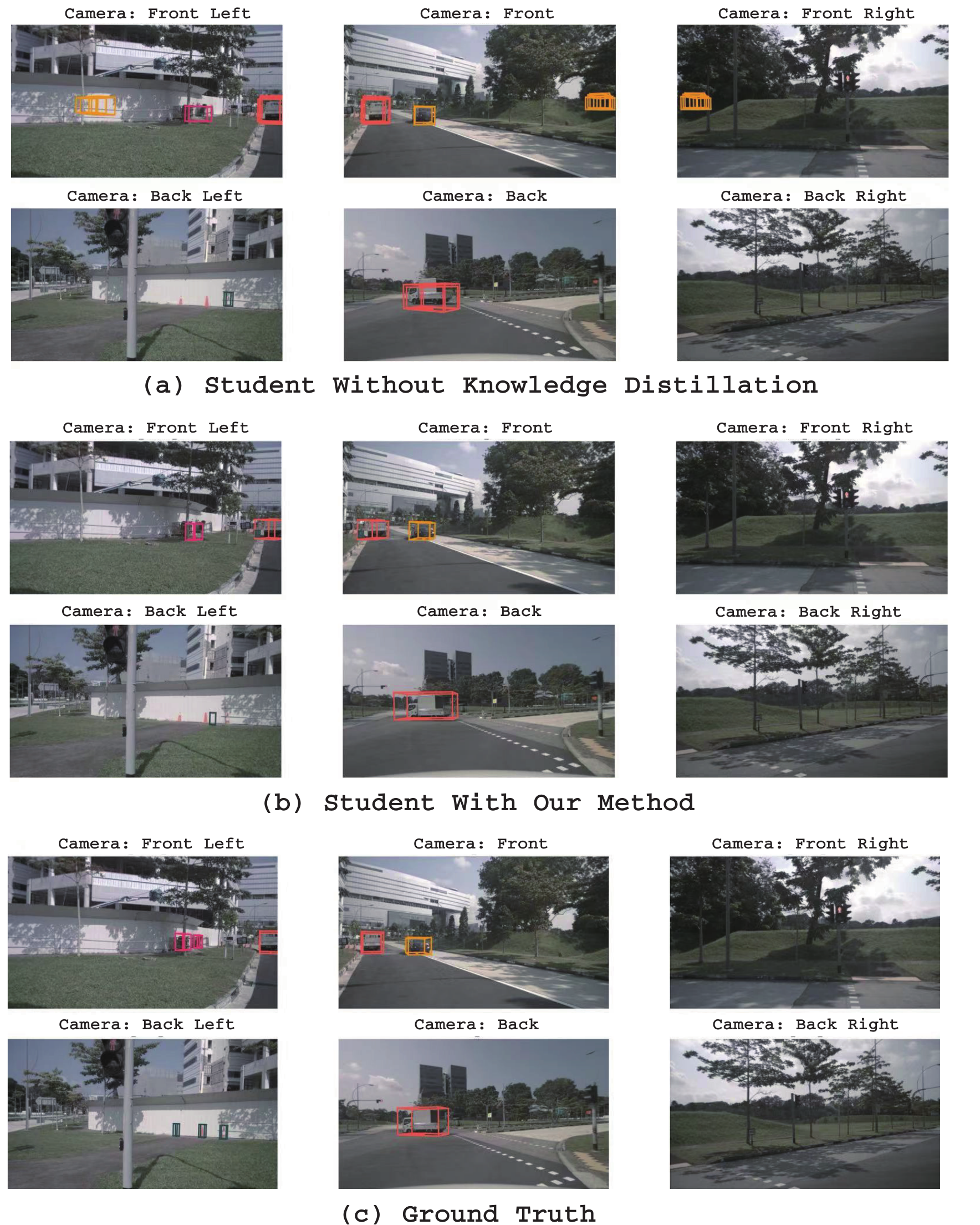}
    \vspace{-0.3cm}
    \caption{Visualization of detection results of images in different views. The color of boxes indicates the corresponding categories.}
    \vspace{-0.2cm}
    \label{fig:detection_vis_camera}
\end{figure}
\subsection{Visualization}
\begin{figure}
    \centering
    \includegraphics[width=\linewidth]{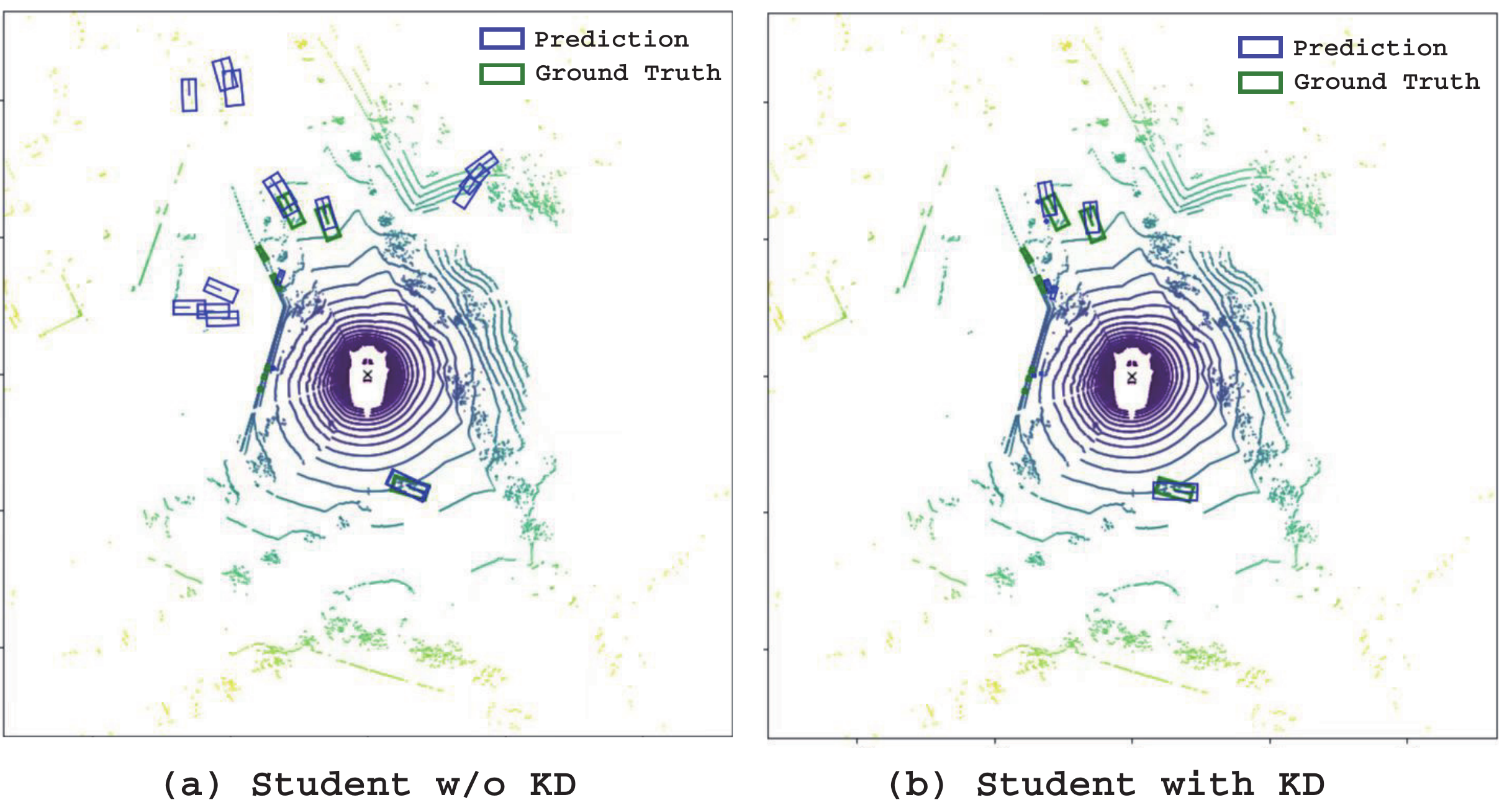}
    \caption{Visualization of detection results in BEV. Boxes in green and blue indicate the ground truth and the prediction results.}
    \vspace{-1em}
    \label{fig:bev_vis}
\end{figure}

\paragraph{Detection Results}
Figure~\ref{fig:detection_vis_camera} and Figure~\ref{fig:bev_vis} 
visualize the detection results of the student detector trained without and with our method from the perspective of different camera views and bird-eye-view, respectively. Note that the used student detector has 6.4 FPS and 40.45M parameters.
It is observed that the student trained by our method produces impressive results which are similar to the ground truth. In contrast, the student trained without knowledge distillation generates incorrect predictions in the cameras of the front-left view, the front-view, and the front-right view. As shown in their BEV visualization, the mistakes made by the student trained without knowledge distillation have a relatively long distance from the car, indicating the student trained without knowledge distillation is unable to detect the faraway objects while our method can address this problem.

\begin{figure}
    \centering
    \includegraphics[width=8cm]{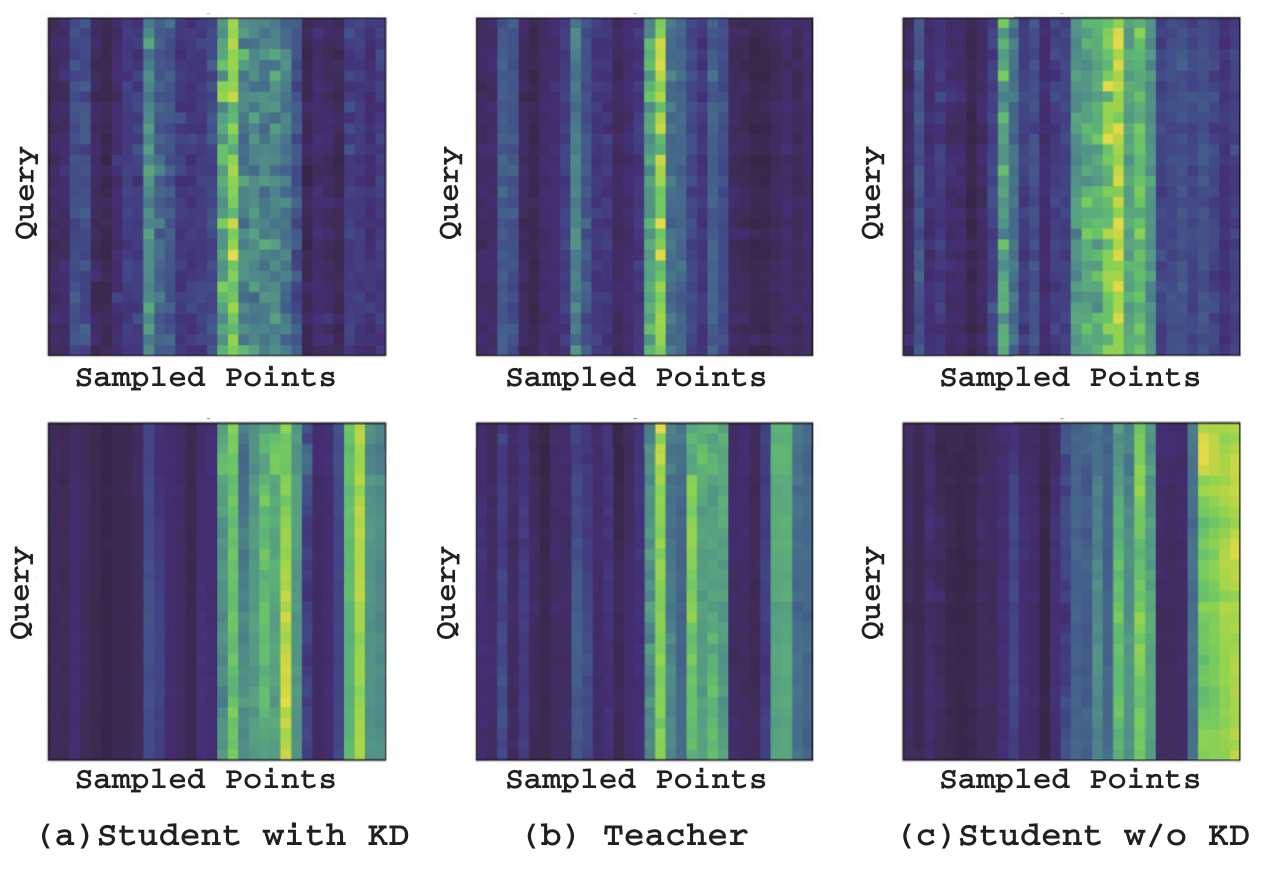}
    \vspace{-0.3cm}
    \caption{Visualization of attention weights in temporal cross-attention from the teacher, the student trained with and without KD. A lighter pixel indicates a higher value.}
    \vspace{-0.2cm}
    \label{fig:bev_atten}
\end{figure}

\vspace{-0.30cm}
\paragraph{Attention Weights}
The proposed spatial-temporal distillation enables the student to learn teacher knowledge on information fusion by training it to mimic the attention weights in temporal self-attention and spatial cross-attention.
Figure~\ref{fig:bev_atten} gives the visualization results of attention weights from the teacher, the student trained with knowledge distillation and the student trained without knowledge distillation. It is observed that: (i) Compared with the student detectors, the attention weights from the teacher tend to concentrate more on several sampled points, indicating the teacher detector is able to leverage the information from certain images. (ii) Compared with the student trained without knowledge distillation, the attention weights from the student trained with knowledge distillation are more similar to the attention weights of the teacher, indicating that the spatial-temporal distillation successfully enables the student to approach the teacher detector.

\begin{figure}
    \centering
    \includegraphics[width=\linewidth]{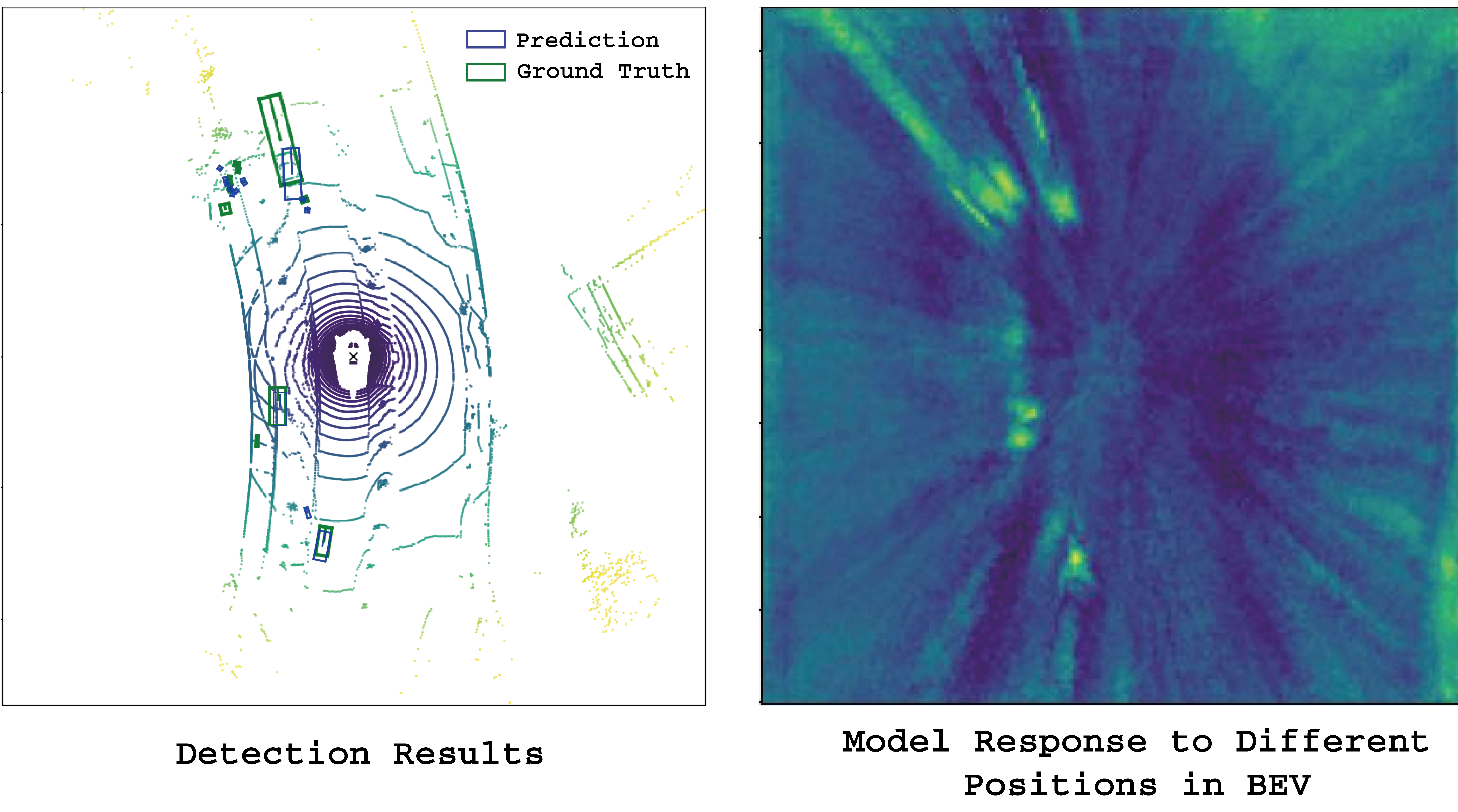}
    \vspace{-0.6cm}
    \caption{Visualization of model response to different positions from Birds'-Eye-View and the corresponding detection results.}
    \vspace{-0.2cm}
    \label{fig:response}
\end{figure}


\vspace{-0.30cm}
\paragraph{BEV Response}
Figure~\ref{fig:response} shows the BEV response and the corresponding detection results from the student detector. Note that a lighter pixel in BEV response map indicates the detector has a higher response.
It is observed that the detector tends to show a higher response in the position where objects exist, indicating that BEV response contains valuable semantic information about the localization of objects. Hence, the proposed BEV response distillation can improve the ability of localization of the student detector by training it to imitate the BEV response from its teacher. 

\section{Conclusion}
\label{sec:conclusion}
Most advanced multi-view 3D detectors suffer from low inference efficiency, which has limited their applications in edge devices. To address this problem, we propose a series of knowledge distillation methods to achieve model compression, which includes (1) spatial-temporal distillation which allows the student to learn how to fuse information from different timestamps and views (2) BEV response distillation which enables the student to learn the localization-aware knowledge, and (3) a weight-inheriting scheme which fixes the BEV queries and positional encoding to guarantee that students and teachers have the same inputs. Extensive comparison experiments with 8 previous methods and sufficient ablation studies demonstrate the significant performance of our method in three different student-teacher settings. On average, 2.16 mAP and 2.27 NDS improvements can be observed on the nuScenes dataset. Besides, detailed ablation studies and visualization have demonstrated the performance of our method from different perspectives. We hope that this paper may promote more research on efficient multi-view 3D detection.

\vspace{-0.4cm}
\paragraph{Limitations}
We tailor our proposed framework for transformer-based multi-camera BEV detection models.
Adaptations must be made to accommodate the popular lift-splat-shoot style BEV detection models\cite{huang2021bevdet,huang2022bevdet4d,bevdepth}.

{\small
\bibliographystyle{ieee_fullname}
\bibliography{egbib}
}

\end{document}